\documentclass[conference]{IEEEtran}
\IEEEoverridecommandlockouts
\usepackage{cite}
\usepackage{amsmath,amssymb,amsfonts}
\usepackage{algorithmic}
\usepackage{graphicx}
\usepackage{textcomp}
\usepackage{xcolor}
\usepackage{stfloats}
\usepackage{url}
\usepackage{balance}
\def\BibTeX{{\rm B\kern-.05em{\sc i\kern-.025em b}\kern-.08em
    T\kern-.1667em\lower.7ex\hbox{E}\kern-.125emX}}
\begin{document}

\title{Lane Change Decision-making through Deep Reinforcement Learning with Rule-based Constraints
\thanks{This work is supported by the National Natural Science Foundation of China (NSFC) under Grants No. 61803371, No. 61573353, No. 61533017 and No. 61873268. This work is also supported by Noah's Ark Lab, Huawei Technologies.
}
}

\author{
\IEEEauthorblockN{Junjie Wang \IEEEauthorrefmark{1}\IEEEauthorrefmark{2}, Qichao Zhang \IEEEauthorrefmark{1}\IEEEauthorrefmark{2}, Dongbin Zhao \IEEEauthorrefmark{1}\IEEEauthorrefmark{2}, Yaran Chen \IEEEauthorrefmark{1}\IEEEauthorrefmark{2}}

\IEEEauthorblockA{\IEEEauthorrefmark{1}The State Key Laboratory of Management and Control for Complex Systems, Institute of Automation, \\Chinese Academy of Sciences, Beijing, China}
\IEEEauthorblockA{\IEEEauthorrefmark{2}University of Chinese Academy of Sciences, Beijing, China \\
Email: \{wangjunjie2017, zhangqichao2014, dongbin.zhao, chenyaran2013\}@ia.ac.cn}
}

\maketitle

\begin{abstract}
Autonomous driving decision-making is a great challenge due to the complexity and uncertainty of the traffic environment.
Combined with the rule-based constraints, a Deep Q-Network (DQN) based method is applied for autonomous driving lane change decision-making task in this study.
Through the combination of high-level lateral decision-making and low-level rule-based trajectory modification, a safe and efficient lane change behavior can be achieved. With the setting of our state representation and reward function, the trained agent is able to take appropriate actions in a real-world-like simulator. The generated policy is evaluated on the simulator for 10 times, and the results demonstrate that the proposed rule-based DQN  method outperforms the rule-based approach and the DQN method.
\end{abstract}


\section{Introduction}
\subsection{Background}
Autonomous driving is widely regarded as one of the most important ways to alter transportation systems, for example, it has the potential to eliminate traffic accidents that are almost entirely caused by human improper operations \cite{fagnant2015preparing}.
Typically, an autonomous driving system consists of four modules: an environment perception module, a decision-making module, a control module, and an actuator mechanism module \cite{li2018reinforcement}. The task of the decision-making module is to make an appropriate driving decision according to the sensor information from the perception module, and then plan a drivable path  to the control module.
Reasonable decision-making in a variety of complex environments is a great challenge, and it is impossible to enumerate coping strategies in various situations. Therefore, a method that can learn a suitable behavior from its own experiences would be desirable.
Over the past few years, deep reinforcement learning (DRL) has made breakthroughs in many fields of artificial intelligence (AI), such as games \cite{mnih2013playing,mnih2015human,silver2016mastering,silver2017mastering} and autonomous driving \cite{zhao2017model,sallab2017deep}.
Deep learning has a strong ability in learning of representations and generalization of matching patterns from the raw data like images and videos \cite{lecun2015deep}, while reinforcement learning has a good capacity of decision-making based on low-dimension features \cite{zhao2016review}. Therefore, DRL algorithms are very effective in tasks requiring both feature representation and policy learning, e.g., autonomous driving lane change decision-making.

\subsection{Related Work}
To the best of our knowledge, the methods for the autonomous driving decision-making task can be classified into two main categories: rule-based and learning-based. While  rule-based methods have achieved some successful applications, the learning-based approaches have also demonstrated their capability in recent years.

Most traditional techniques are based on manually designed rules and typically rely on state machines to switch between predefined decision behaviors \cite{schwarting2018planning}. For instance, the ``Boss'' developed by CMU \cite{urmson2008autonomous} determined the triggering of lane changing behavior according to certain rules and a preset threshold. Similarly, researchers at Stanford \cite{montemerlo2008junior} used a series of reward designs and a finite state machine to determine the trajectory and steering instructions of driverless cars. Nevertheless, how to make a reliable decision is still a problem for the rule-based methods owing to  their poor ability to generalize to unknown situations \cite{schwarting2018planning}.

 As a core technology of AI, learning-based approaches can provide more advanced and safe decision-making algorithms for  autonomous driving. In 1989, ALVINN (Autonomous Land Vehicle in a Neural Network) \cite{pomerleau1989alvinn} started an early attempt on end-to-end driving by training a neural network to map from camera images to the steering angle. Since then, learning-based autonomous driving decision-making algorithms have been studied extensively. Recently, NVIDIA researchers \cite{bojarski2016end} trained a deep convolutional neural network (CNN) to map raw images of three front-facing cameras directly to the steering command in a real vehicle. During training, the images were fed into a CNN after random shift and rotation. Then the network outputted a steering command which was compared to the desired command. And the error between the two steering angles was used to adjust the network parameters through back propagation. The trained model was able to handle the task of lane and road following such as driving on a gravel road.

In addition to end-to-end supervised learning, reinforcement learning has also been widely used in the autonomous driving decision-making  task since it has the capability of dealing with time-sequential problems and seeking optimal policies for long-term objectives. Wolf et al. \cite{wolf2017learning} presented a method for teaching an agent to drive a car in a simulation environment by using DQN. The input of the network was the front-facing camera image with a shape of 48\(\times\)27, and the action space, i.e. the network output, was discrete---5 actions were defined and each corresponding to a different steering angle. The reward function depended on the distance from the lane center and some relative items (such as the error of the angle between the vehicle and the center line). Hoel et al. \cite{hoel2018automated} also used DQN to deal with the problem of vehicle speed and lane change decision-making in a simulation environment. Different from previous work mentioned above, the Q-network input in \cite{hoel2018automated} was defined as a one-dimensional vector of the relative position, speed, and lane of surrounding vehicles, rather than front-facing images. Two different agents were defined with the discrete action space. The first agent considered only the lateral lane change control, while the second considered both the lane change and speed control. The result showed  that the second agent outperforms the former when CNNs were both utilized. However, how to guarantee the decisions' safety of the learning-based method should be further considered. 

\subsection{Overview}
This paper presents the details of a DQN algorithm with the rule-based constraints for autonomous driving lane change decision-making in a real-world-like simulation environment.
The rest of this paper is constructed as follows.
The preliminaries and methodology are introduced in Sect.~2. Next, the used simulator and our MDP formulation are described in Sect.~3.  In Sect.~4, we first give an overview of our simulation framework and then explain the setting of relevant parameters and training details. Finally, the discussion and conclusions are given in Sect.~5.

\section{Methodology}
\subsection{Markov Decision Process and Reinforcement Learning}
A Markov decision process (MDP) provides a mathematical framework for modeling decision-making in situations where outcomes are partly random and partly under the control of a decision maker. The policy of an MDP has the Markov property, which means that the conditional probability distribution of the future state of a random process depends entirely on the current state when the present state and all the past states are given \cite{sutton2018reinforcement}.

An MDP is a 4-tuple \(M = \left\langle {S,A,{P_{sa}},R} \right\rangle\) \cite{sutton2018reinforcement}, where

\begin{itemize}
\item \(S\) is a set of states, and \(s_i \in S\) is the state in time step \(i\).
\item \(A\) is a set of actions, and \(a_i \in A\) is the action in time step \(i\).
\item \(P_{sa}\) is the probability that action \(a \in A\) in current state \(s \in S\) will lead to next state. In particular, \(p(s'|s,a)\) is the probability that action \(a\) in current state \(s\) will lead to state \(s'\).
\item \(R\) is the reward function. Particularly, \(r(s'|s,a)\) is the expected immediate reward received after transitioning from state \(s\) to state \(s'\), due to action \(a\).
\end{itemize}

The agent interacts with the environment at each of a sequence of discrete time steps, \(t=0,1,2,...\). At each time step \(t\), the agent perceives the current state \(s_t \in S\) of the environment, and then it selects a doable action \(a_t \in A\) based on the state and executes the action \(a_t\). After the action carries out in the environment, the agent receives a numerical reward \(r_t \in R\) and the next state \(s_{t+1}\).

The goal of reinforcement learning is to learn an optimal policy which maps from environmental states to agent's actions by maximizing the cumulative reward of actions taken from the environment \cite{sutton2018reinforcement}. The policy of time step \(t\) denoted by \(\pi_t\) maps from environmental states to probabilities of selecting each possible action, and \(\pi_t(a|s)\) is the probability that \(a_t=a\) if \(s_t=s\). To obtain the optimal policy \(\pi^*\), reinforcement learning typically uses a method of maximizing a total expected return \(G\) that is a cumulative sum of immediate rewards \(r\) received over the long run. At time step \(t\), \(G_t\) is defined as
\begin{equation}
{G_t} = \sum\limits_{k = 0}^\infty {{\gamma ^k}{r_{t + k}}} \label{G},
\end{equation}
where \(\gamma \in (0,1]\) is a discount factor.

\subsection{Q-learning and Deep Q-Network}
Q-learning is an off-policy TD control algorithm in reinforcement learning \cite{watkins1992q}. Basically, a state-action value function \(Q^\pi(s,a)\) for policy \(\pi\) is defined as
\begin{equation}
{Q^\pi}(s,a) = {\mathbb{E}_\pi}\left[{\left.{\sum\limits_{k = 0}^\infty  {{\gamma ^k}{r_{t+k}}}}\right|{s_t} = s,{a_t} = a}\right] \label{Q}.
\end{equation}
It is used to evaluate the particular policy \(\pi\), and the Q-learning algorithm tries to find the optimal state-action value function \(Q^*(s,a)\) defined as
\begin{equation}
{Q^*}(s,a) = \mathop {\max }\limits_\pi{Q^\pi(s,a)}, \label{Qstar}
\end{equation}
which corresponding to the optimal policy \(\pi^*\) \cite{sutton2018reinforcement}.

The value function \(Q^*(s,a)\) follows the Bellman optimality equation
\begin{equation}
{Q^*}(s,a) = \mathbb{E}\left[ {\left. {r + \gamma \mathop {\max }\limits_{a' \in A} {Q^ * }(s',a')} \right|s,a} \right]. \label{Bellman}
\end{equation}
Based on the optimal value function \(Q^*(s,a)\), we can determine the optimal policy by finding one or more actions that maximize the value function at each state\cite{sutton2018reinforcement}.

When the states are discrete and finite, the Q-function can be easily formulated in a tabular form. But in many practical applications, for example, lane change decision-making task, the state space of them is very large or even continuous, using the Q-learning algorithm will lead to dimension disaster. Therefore, tabular Q-learning algorithm does not applicable to the learning problem of continuous state space and continuous action space. 

DQN algorithm proposed in \cite{mnih2013playing} can be used to handle the problem of dimension disaster. In the DQN algorithm, multi-layer neural network (i.e. the Q-network) is used to approximate the value function. The approximate value function is denoted by \(Q(s,a;\theta_i)\), in which \(\theta_i\) are the weights of the Q-network at iteration \(i\). The agent's experiences \(e_t=(s_t,a_t,r_t,s_{t+1})\) at each time step \(t\) stored in a data set \(D_t=\{e_1,e_2,\cdots,e_t\}\) are used to train the Q-network. At iteration \(i\), a mini-batch of experiences \(M\) are randomly sampled from \(D_t\) to update the parameters \(\theta_i\) with the loss function
\begin{equation}
{L_i}({\theta _i}) = {\mathbb{E}_M}\left[ {{{\left( {r + \gamma \mathop {\max }\limits_{a'} Q(s',a';\theta _i^ - ) \\
- Q(s,a;{\theta _i})} \right)}^2}} \right], \label{loss}
\end{equation}
where \(\theta _i^ -\) are the network parameters used to compute the target at iteration \(i\). The target network parameters \(\theta _i^ -\) are only updated using the Q-network parameters \(\theta _i\) every \(C\) steps. For computational convenience, the stochastic gradient descent method is commonly used to optimize the loss function.

\section{Problem Formulation}
\subsection{Simulator}
\begin{figure}[htbp]
\centerline{\includegraphics[width=2in]{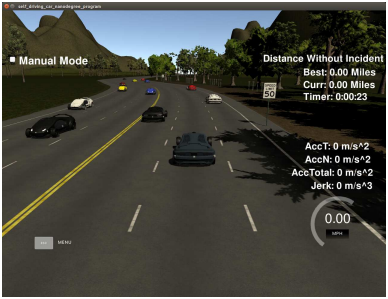}}
\caption{Screenshot of the simulator}
\label{env}
\end{figure}

In this study, we utilize a three-lane highway traffic simulator which is proposed by one of Udacity self-driving car projects for our simulation, see \url{https://github.com/udacity/CarND-Path-Planning-Project}. A screenshot of the simulator is illustrated in Fig.~\ref{env}.

Only three lanes on the right side of the road are considered in this paper. The dark car in the middle lane shown in Fig.~\ref{env} is controlled by our algorithm, namely, the self-driving car, and there are up to 12 interference cars in the simulation environment.
The localization, speed and sensor fusion data (include other cars' localization and speed) of the self-driving car are provided by the simulator. There is also a sparse map list of waypoints around the highway.
The maximum speed of the car cannot exceed 50 MPH. The car should try to drive as fast as possible  and avoid collisions with other cars.  It can try to change lanes safely when there is a front car blocking. Also, the car should not experience total acceleration over 10 m/s\(^2\) and jerk greater than 10 m/s\(^3\), and it should be able to make one complete loop around the 6946m highway.

\subsection{MDP Framework}
We regard the autonomous driving decision-making process as an MDP, in which the self-driving car interacts with the environment including surrounding vehicles and lanes. We define the state, action, and reward function for driving policy learning as follows.

1) State

As mentioned, the information offered by the simulator includes the position and speed of the self-driving car and other cars. Specifically, it includes cars' \(x\), \(y\) position in map coordinates and \(s\), \(d\) position in frenet coordinates, the self-driving car's yaw angle in the map and speed in MPH, and the other cars' \(x\), \(y\) speed in m/s. In this paper, the \(s\), \(d\) positions and speeds of each vehicle are used to represent the state of the simulation environment. We unite the unit of speed to MPH and convert traffic conditions to a grid form. 

\begin{figure}[htbp]
\centerline{\includegraphics[width=3.3in]{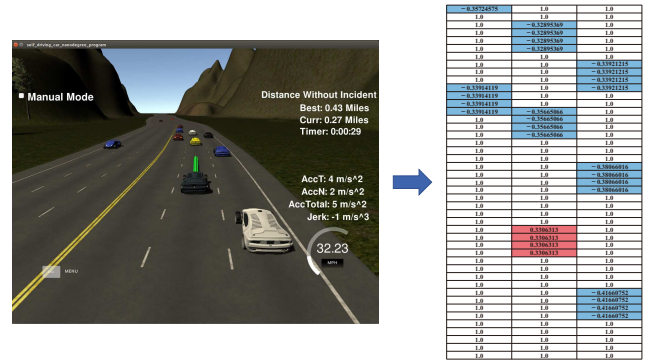}}
\caption{The state representation}
\label{state}
\end{figure}

As shown in Fig.~\ref{state}, we use a \(45\times3\) matrix as the state representation in this study. The whole matrix corresponds to traffic situation within the range of 60 meters in front and 30 meters behind the self-driving car in the three lanes on the right side of the road. Since cars are approximately 5.5 meters in length and each row in the matrix spans 2 meters longitudinal, one car can occupy 4 cells in extreme cases. Therefore, we fill the 4 cells corresponding an individual car with the car's normalized speed (within \([0,0.5)\)), and the normalized values of speed are positive to self-driving car (the red block shown in Fig.~\ref{state}) while other cars' are negative (the blue blocks shown in Fig.~\ref{state}). Where there is no car, the corresponding cells are filled with a uniform value (set to \(1\) in this study).

2) Action

\begin{table}[!htb]
  \centering
  \caption{The set of actions}
  \label{tab1}
  \begin{tabular}{c|c}
    \hline \hline
    Decision          & \multicolumn{1}{c}{Action} \\ \hline
    \({a_0}\)         & Stay in current lane \\
    \({a_1}\)         & Change lanes to the left \\
    \({a_2}\)         & Change lanes to the right \\
    \hline \hline
  \end{tabular}
\end{table}

When it comes to an autonomous-driving-decision-making problem, both longitudinal speed and lateral lane change decisions need to be considered. In this paper, we focus on lane change decision-making with reinforcement learning, and longitudinal speed control is implemented in a rule-based way.
We define three actions in this study, and the action space (given in Table~\ref{tab1}) of the agent is then discrete.

3) Reward

Typically, safety and efficiency are the main concerns for a lane change process. In terms of safety, autonomous cars should be able to avoid collisions while driving, and they also need to travel in prescribed lanes.
In order to force the autonomous car to drive only in the prescribed lanes, we limit it to only drive in the three lanes on the right side of the road. That means when a decision to change lanes to the left (\({a_1}\) in Table~\ref{tab1}) is made while the car is on the left lane (or \({a_2}\) on the right lane), it stays in current lane, but a penalizing reward \(r_{ch1}=-5\) will be given to the agent. To avoid the collisions, a large penalizing reward \(r_{co}=-10\) is then given to the agent if a lane changing decision results in the self-driving car collides.

For efficiency, the self-driving car should try to meet the requirements of driving as fast as possible without exceeding the maximum speed limit, and not changing lanes too often. In order to make it travel faster, we define the following reward according to the speed of the car

\begin{equation}
{r_v} = \lambda  \cdot \left( {v - v_{ref}} \right), \label{rv}
\end{equation}
where \(v\) denotes the car's average speed in MPH within one decision period since last decision-making, and \(v_{ref}\) is reference speed while \(\lambda\) is a normalizing coefficient. Considering the speed of the self-driving car is between 0 and 50 MPH (less than 50 MPH), we set \(v_{ref}=25\) and \(\lambda=0.04\), so that the value of reward is normalized to the range of \([ - 1,1)\). To avoid the meaningless lane change behavior, a negative reward \(r_{ch2}=-3\) is given if the car chooses to change lanes while there is no car in front of it. In addition, a small negative reward \(r_{ch3}=-0.3\) is added to the basic reward \(r_v\) when lane change happens without collision.
In general, our reward function goes as
\begin{equation}
{r}=
\begin{cases}
{r_{co}}& \text{collision happens}\\
{r_{ch1}}& \text{illegal lane change}\\
{r_{ch2}}& \text{invalid lane change}\\
{\lambda  \cdot \left( {v - v_{ref}} \right)+r_{ch3}}& \text{legal lane change}\\
{\lambda  \cdot \left( {v - v_{ref}} \right)}& \text{normal drive}
\end{cases}.
\label{r}
\end{equation}

4) Rule-based constraints

To ensure the absolute security of the lane changing behavior, we add the rule-based constraints based on the planning trajectories and the others' predicted trajectories. When the high-level decision maker determines an action, the low-level controller can predict the trajectories of the ego car and the surrounding cars in the intend lane based on this action. The trajectory of the ego car is the planning driving trajectory by path planning. And surrounding vehicles' trajectories are made by assuming they maintain the current speed and keep in the current lane. If there is a moment when the distance between the predicted trajectory of the ego car and that of surrounding vehicles is less than the predefined threshold value, the decision made by the high-level decision maker is potentially dangerous. This action will be canceled and the ego car stays in the current lane. In other words, when the high-level decision maker makes a dangerous decision, the low-level decision maker will modify it according to the surrounding environment to guarantee safety.

\section{Simulation and Results}
\subsection{Simulation Setup}
\begin{figure*}[htbp]
\centerline{\includegraphics[width=6.5in]{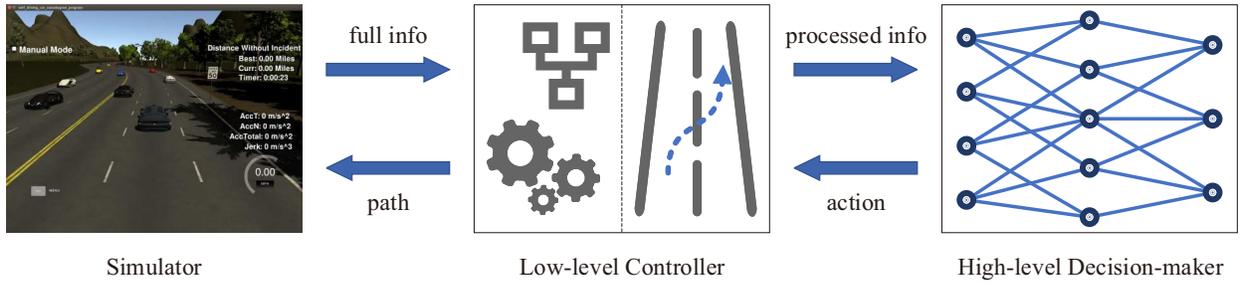}}
\caption{Simulation framework}
\label{sim}
\end{figure*}

The implementation of our simulation experiment consists of three parts, as shown in Fig.~\ref{sim}. The first part is the simulator (introduced in Section 3), which outputs the environment information and receives a specified path. The second part is controller and planner, which is responsible for controlling speed and planning path. The third part is the DQN algorithm, which is in charge of high-level lane change decision-making.

In this study, we aim at deep reinforcement learning based lateral lane change decision-making. In Fig.~\ref{sim}, the middle block representing low-level controller shows the rules-based speed controller in the left half and the path planner in the right half while data processing part is omitted. The longitudinal speed control is implemented by using a rule-based approach, while path planning is implemented by using spline interpolation according to the provided waypoints combined with the potential target points corresponding to the lane change decision result. Simultaneously, the controller also acts as a lower-level modifier to revise the higher-level decisions.

The interaction between the three parts in Fig.~\ref{sim} is as follows: the simulator provides environmental information including vehicle speed and position to the controller. After preliminary processing (such as calculating reward value) of the information, it transmits the processed information to the decision maker. The decision maker first obtains the grid form state based on this information, and then DQN computes an action according to the current state and returns it to the planner. The planner finally outputs an appropriate path to the simulator according to the action combined with speed control. The following presentation focuses on the decision-making part.

\subsection{Training}

1) Architecture of artificial neural network

\begin{figure}[htbp]
\centerline{\includegraphics[width=3in]{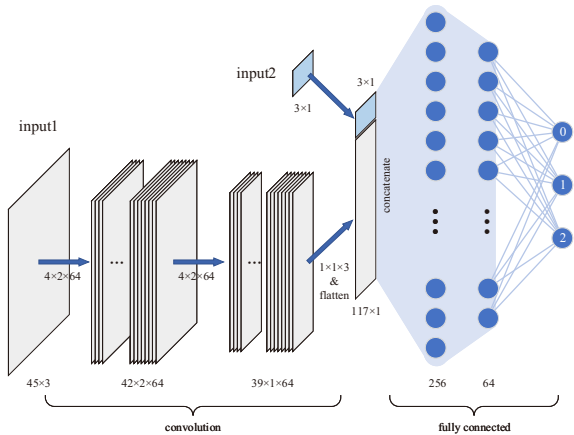}}
\caption{Artificial neural network architecture}
\label{cnn}
\end{figure}

In recent years, CNNs that inspired by biological processes \cite{matsugu2003subject} have become one of the research focuses of many scientific fields, especially in the field of machine learning (ML). Since CNNs avoid the complex pre-processing of images and can receive inputs from the raw images directly, they have been widely used in many vision-related domains.
Benefits from locally connected and shared-weights architecture and translation invariance characteristics, the amount of parameters in the network is vastly reduced. In this study, a CNN architecture is used as our Q-network (as shown in Fig.~\ref{cnn}). The input2 shown in Fig.~\ref{cnn} is a 3\(\times\)1 one-dimensional vector, defined as
\[\begin{array}{l}
{s_1}\  \text{Normalized ego vehicle speed}\\
{s_2}
\begin{cases}
1& \text{if there is a lane to the left}\\
0& \text{otherwise}
\end{cases}.\\
{s_3}
\begin{cases}
1& \text{if there is a lane to the right}\\
0& \text{otherwise}
\end{cases}
\end{array}\]

2) Training details

In terms of reinforcement learning, the experience is of great importance for an agent to learn a good policy. As mentioned in Sec.~2, the agent's experiences are stored in an experience pool, and during training, random samples from it are used to update the network parameters. Thus, the experience of the agent determines its performance. By applying experience replay, the correlations between the samples are broken and the variance of the updates is reduced \cite{mnih2015human}. In order to enable the agent to learn more effective policies, we divide the experience pool into two parts, one is to store the experience of keeping in lane action and the other is to store the experience of lane changing action. Then the network is updated by sampling half batch size experiences from each of these two parts.

Since the appropriate training samples are needed, the agent must face the balance between exploration and exploitation. On the one hand, the agent need explores more and discovers some unknown states and actions because some of them may bring a higher reward. On the other hand, it is necessary for the agent to use the knowledge it has learned to choose actions. This will ensure that the experience is effectively used and the convergence of the learning process is accelerated. 
The trade-off between exploration and exploitation in this study is handled by following an \(\varepsilon\)-greedy policy. The main idea of the \(\varepsilon\)-greedy algorithm is to randomly select an action from the action space with the probability of \(\varepsilon\), and to select the current optimal action according to the greedy method with the probability of \(1-\varepsilon\).
When \(\varepsilon=0\), only the current optimal action is selected, and when \(\varepsilon=1\), the action is completely randomly selected. In this study, \(\varepsilon\) is not a constant value, but it will slowly decline with the iteration goes, that is,

\begin{equation}
\varepsilon  = \max ({\varepsilon _0} \cdot {\lambda _{decay}}^{step},{\varepsilon _{\min }}).
\label{epsilon}
\end{equation}
We set \({\varepsilon _0} = 1.0\), \({\lambda _{decay}} = 0.99985\) and \({\varepsilon _{\min }} = 0.03\).

\subsection{Results and Evaluation}
\begin{figure}[htbp]
\centerline{\includegraphics[width=3.3in]{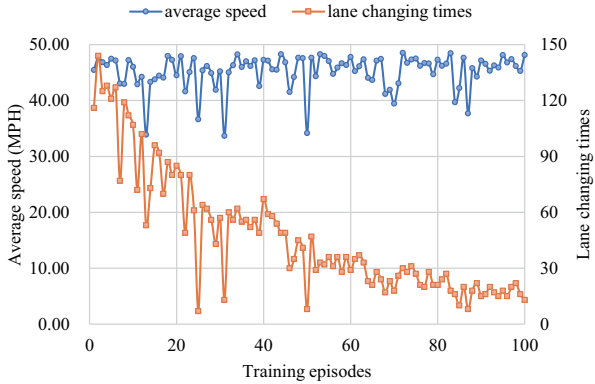}}
\caption{The agent's performance during training}
\label{training}
\end{figure}

The simulator runs at a real-world time, and it takes about 6 minutes for the ego car to complete a circle in the simulation environment, which is seen as an episode of our training process. And our training process is composed of 100 episodes. When one episode is completed, that is, the ego vehicle returns to the initial point, we restart the simulator for the next episode of training. The average speed and the number of lane changing times of one specific episode during training for the proposed rule-base DQN are shown in Fig.~\ref{training}.

As can be seen from Fig.~\ref{training}, the agent generally achieves an ideal behavior. As the training progresses, the average speed of the ego vehicle increases in general while the frequency of lane change reduces significantly. These trends indicate that the agent gradually learned effective lane changing policy during the training process, and achieved faster speed with fewer lane changing behaviors. At the same time, it can be seen that a smaller number of lane changing times usually corresponds to a slower average speed, which may be because the ego car is blocked and cannot change lanes.

We evaluate the performance of the trained rule-based DQN agent by comparing its average speed and average lane changing times to other different approaches, i.e. a random policy with collision avoidance constraints, a rule-based policy and a DQN-based policy. For each  method, we test 10 times in the simulation environment and then calculate its average speed, the average number of lane changing times and the safety rate.  The agent chooses actions randomly when it is following the random-action policy. For the rule-based policy, the agent makes a lane change decision when the distance between the ego car and the very front car in its lane is less than 20 meters, and simultaneously the front car in the neighbor lane is further to the ego car. It has a tendency to switch to the left when both sides are available. All four methods use the same longitudinal controller and low-level corrector.
The results are summarized in Table~\ref{tab2}, where \(\bar v\) denotes the average speed and \(c_{ch}\) denotes the number of lane changing times of one test episode. At the same time, the safety rate is the ratio of the number of test episodes without collisions to the total number of test episodes. In terms of both average \(\bar v\) and average \(c_{ch}\), the proposed rule-based DQN is superior to others. In fact,  DQN methods can improve the average speed compared to the rule-based methods, and lane changing occurs less frequently for the proposed algorithm. Note that the rule-based DQN policy has a higher safety rate than the DQN-based policy, which can guarantee the safety of the lane changing decision. This suggests that our approach achieves a more efficient and safe policy than the others.

\begin{table}[!htb]
  \centering
  \caption{Comparison between different approaches}
  \label{tab2}
  \begin{tabular}{l|c|c|c}
    \hline \hline
                            & avg \(\bar v\) (MPH)  & avg \(c_{ch}\)    & safety rate\\ \hline
    random-action policy    & 44.59                 & 152.60            & 0.6\\
    rule-based policy       & 45.22                 & 8.40              & 0.6\\
    DQN-based policy        & 46.16                 & 37.40             & 0.2\\
    rule-based DQN policy   & 46.99                 & 8.80              & 0.8\\
    \hline \hline
  \end{tabular}
\end{table}

\balance

\section{Conclusion and Future Work}
In this paper, we present the details of deep reinforcement learning algorithm with rule-based constraints to handle the problem of high-level lane change decision making. First, a real-world-like simulation environment is used. Different from other simulation environments, it needs an explicit driving path, and it covers both dense and sparse traffic conditions. Second, through the combination of high-level lateral decision-making and rule-based low-level longitudinal control and trajectory modification, safe and efficient driving behavior can be achieved. Third, the rule-based DQN is used to map the representation of the surrounding environment to the lateral decision. Due to the setting of our state representation and reward function, the trained agent is able to take appropriate actions in different scenarios. The generated policy is evaluated on the simulator for 10 times. The results demonstrate the effectiveness of the proposed method.

There still exist some shortcomings in this work that need to be improved in the future. For example, the historical trajectory information of vehicles can be taken into account as part of the state representation. And also, more advanced deep reinforcement learning algorithms can be adopted.


\end{document}